\PassOptionsToPackage{numbers}{natbib}
\documentclass{ijisa}

\usepackage[utf8]{inputenc}
\usepackage[english]{babel}
\usepackage{newtxtext,newtxmath}
\usepackage{graphicx}
\usepackage{amsmath,amssymb}
\usepackage{subcaption}
\usepackage{booktabs}
\usepackage{siunitx}
\usepackage{stfloats}
\usepackage{xcolor}
\usepackage{enumitem}
\usepackage{natbib}
\usepackage{hyperref}
\hypersetup{
  colorlinks = true,
  urlcolor   = blue,
  linkcolor  = blue,
  citecolor  = blue
}
\setlist{nolistsep,leftmargin=*}
\title{Chance-Constrained Inspection Viewpoint Sequencing for On-Orbit Servicing via Safe Reinforcement Learning\\}

\author[1]{Chlo\'{e} Carer}
\author[1,*]{Adam Abdin}

\affil[1]{Laboratoire G\'{e}nie Industriel, CentraleSup\'{e}lec, Universit\'{e} Paris-Saclay, Gif-sur-Yvette, France.\vspace*{.2cm}}
\affil[*]{Corresponding author. e-mail: adam.abdin@centralesupelec.fr}

\abstract{
Autonomous inspection of a target spacecraft is typically the initial step in on-orbit servicing. Planning close-proximity observation viewpoints involves a complex combinatorial problem constrained by coverage, fuel, and safety considerations. This paper models the viewpoint selection as a Constrained Markov Decision Process (CMDP) and applies Proximal Policy Optimization with Lagrangian relaxation to solve it. The system combines a high-level planner, which sequences camera viewpoints on an observation sphere around the target, with a low-level model that assesses each transfer using the Clohessy--Wiltshire (CW) equations for relative motion. The deterministic constraints on single-impulse reachability and $\Delta v$ cap are enforced directly on the action set. Safety margins are affected by navigation and manoeuvre errors, making the actual clearance stochastic. To address this, a chance constraint on the realised trajectory is implemented, with a Lagrangian multiplier governed by PID control to adapt the penalty during training. The approach is tested on a Galileo-class navigation satellite in medium Earth orbit, modelled as a point cloud where visibility from each viewpoint considers field-of-view and line-of-sight occlusion. Across various risk budgets, the trained policy maintains the keep-out violation rate within the specified limit, achieving coverage close to its maximum and staying within the mission $\Delta v$ budget. Compared to the usual practice of applying a fixed margin to the keep-out radius and planning deterministically, the chance-constrained policy attains less than half the violation rate at comparable coverage. This makes the collision risk an explicit design parameter, eliminating the need for re-tuning per case. The learned policy selects a viewpoint without the repeated Monte Carlo risk estimates required by a greedy, chance-constrained planner at every decision point.
}


\begin{document}

\maketitle
\abstract
\keywords
\thispagestyle{thefirstpage}

\section*{Nomenclature}
\begin{description}[leftmargin=4.2em,style=nextline,labelwidth=3.6em,font=\normalfont]
\item[CMDP] Constrained Markov Decision Process
\item[CW] Clohessy--Wiltshire
\item[FOV] Field of View
\item[KOS] Keep-Out Sphere
\item[MDP] Markov Decision Process
\item[MEO] Medium Earth Orbit
\item[NBV] Next-Best-View
\item[OOS] On-Orbit Servicing
\item[PID] Proportional--Integral--Derivative
\item[PPO] Proximal Policy Optimization
\item[RL] Reinforcement Learning
\item[RPO] Rendezvous and Proximity Operations
\item[RSO] Resident Space Object
\item[$\Delta v$] Velocity increment (manoeuvre cost)
\item[$\varepsilon$] Keep-out chance-constraint budget
\item[$\lambda$] Lagrange multiplier
\end{description}

\section{Introduction}
\label{section:intro}

On-orbit servicing (OOS), which includes inspection, maintenance, repair, refuelling, and removal of spacecraft already in orbit, is transitioning from isolated flight demonstrations to practical, operational activities. This shift is driven by the need to prolong the lifespan of high-value assets and to address the increasing congestion in Earth's orbital environment \citep{li2019oos,corbin2020osam}. Servicing and active-debris-removal missions usually start with an inspection phase. A chaser spacecraft approaches a resident space object (RSO) and captures images from various viewpoints to assess its shape, orientation, and condition before any physical interaction occurs. The data collected during inspection limits the scope of subsequent operations, making it a critical preparatory step.

We approach autonomous inspection as a sequential decision-making task. The chaser navigates through a series of camera viewpoints around the target, observing the visible and unoccluded surface areas within the sensor's FOV from each position. The goal is to select viewpoints that collectively maximise surface coverage, while ensuring each movement adheres to the spacecraft's propulsion constraints and maintains a safe distance outside the keep-out sphere (KOS) surrounding the target. This creates a constrained sequencing problem that integrates coverage goals with orbital dynamics and safety considerations, which must be solved on board, under limited time and computation.

Reinforcement learning (RL) tackles this problem by creating a reactive policy that directly links the current inspection state to the next manoeuvre, shifting the planning costs to an offline training phase \citep{sutton2018}. Recent studies have employed RL for spacecraft proximity operations and inspections \citep{vanwijk2023inspection,aurand2025inspection,mu2024saferl,bourriez2023cam}. The main challenge, however, is ensuring safety. Operations at close range are inherently uncertain. Relative navigation is imperfect, manoeuvre execution can contain errors, and linearised relative dynamics often ignore external perturbations such as $J_2$, drag, and solar radiation pressure \citep{wakker2015}. Given these uncertainties, whether a commanded transfer remains outside the keep-out sphere is a probabilistic event that the planner can only estimate. Therefore, a deterministic feasibility filter that approves or denies transfers based on their nominal trajectory may be overly optimistic---since the actual trajectory might differ---or excessively conservative, as it might account for worst-case scenarios in terms of coverage and fuel.

Safe reinforcement learning, particularly the constrained Markov decision process (CMDP), offers a formal framework for this context \citep{altman1999cmdp,gu2024saferl}. A CMDP constrains the \emph{expected} value of a cost signal, enabling the formulation of the keep-out requirement as a chance constraint---an upper bound $\varepsilon$ on the probability that a transfer breaches the keep-out sphere. Lagrangian relaxation introduces a multiplier, $\lambda$, which penalises violations within the objective function, thereby allowing a single tunable parameter to determine the operating point of the resultant policy. Prior reinforcement learning research on spacecraft inspection enforces keep-out safety deterministically---either through run-time assurance filters, several of which are based on control barrier functions \citep{vanwijk2024cbf,dunlap2024sixdof,dunlap2023rta}, or through fixed reward penalties---and reports performance at a single safety configuration. When such studies vary a design parameter, as observed in the run-time assurance analysis by \citet{hamilton2023ablation}, the varied parameters pertain to the assurance mechanism, while the tolerated level of risk remains fixed. Chance-constrained reinforcement learning has been applied in other areas of space guidance, such as interplanetary transfer and pinpoint landing \citep{chaudhary2026chance}, but has not been employed for inspection viewpoint sequencing. To our knowledge, the behaviour of a calibrated keep-out risk budget under stochastic relative dynamics has not been examined.

This paper presents the following contributions:
\begin{itemize}
  \item[(i)] We formulate autonomous viewpoint sequencing as a \emph{chance-constrained} Markov Decision Process (CMDP) under stochastic relative dynamics, where safety margins are defined by a collision risk budget, $\mathbb{P}[\text{KOS violation}] \le \varepsilon$, specified on the realised trajectory, with execution and navigation errors contributing to the risk.
  \item[(ii)] We distinguish deterministic reachability from stochastic safety. Single-impulse Clohessy--Wiltshire (CW) reachability, constrained by a per-manoeuvre $\Delta v$ limit, is enforced on the planned impulse through invalid-action masking \citep{huang2022masking}, while the keep-out risk is managed by a constrained policy, weighted by a PID Lagrangian multiplier \citep{stooke2020pid}.
  \item[(iii)] We demonstrate the approach on a Galileo-class navigation satellite in medium Earth orbit (MEO), modelled as a $10{,}000$-point cloud with field of view (FOV) and line-of-sight occlusion, and report the realised violation rate across risk budgets spanning $\varepsilon\in[0.01,0.20]$.
  \item[(iv)] We benchmark our method against random, coverage-greedy, robust-mask, and chance-constrained heuristics under the same stochastic dynamics, and show that the chance-constrained treatment attains less than half the keep-out violation rate of a deterministic keep-out margin at comparable coverage.
\end{itemize}

The remainder of the paper is organised as follows. Section~\ref{section:related} reviews related work on autonomous inspection and coverage planning, RL for spacecraft proximity operations, and safe RL. Section~\ref{section:method} presents the problem formulation, the relative-motion model, and the chance-constrained solution approach. Section~\ref{section:results} reports the experimental results, and Section~\ref{section:conclusions} concludes.

\section{Related Work}
\label{section:related}

\subsection{Autonomous inspection and coverage planning}

Selecting where to look to explore an unknown or partially known structure is a common problem in robotics called \emph{next-best-view} (NBV) and coverage planning. Receding-horizon NBV planners build a sampling tree, choose the branch with the most expected new observations, and then execute only its first move before re-planning. This method is widely used for autonomous 3D exploration and inspection \citep{bircher2016nbv}. In orbital scenarios, inspection has also been approached as a trajectory optimisation problem. \citet{connerney2021inspection} optimise fuel-efficient relative trajectories that meet keep-out and plume-impingement constraints. These techniques require an accurate target model and dynamic data, typically obtained through flight, and the optimised trajectory is computed beforehand, so it does not adapt to uncertainties encountered during flight. Our approach maintains the NBV viewpoint-sequencing framework, learns a reactive policy resilient to stochastic dynamics, and keeps a chance-constrained greedy NBV heuristic as a baseline.

Planning has also been examined beyond the scope of individual rendezvous optimisation. \citet{abdin2025spaceops} discusses the application of operations research methodologies for the strategic scheduling of servicing missions, while \citet{nikitits2026adr} propose active debris removal campaigns using deep reinforcement learning, incorporating an explicit consideration of mission risk. These formulations prioritise target sequences across an entire campaign, whereas our approach sequences viewpoints within a single inspection, where risk is influenced by the relative dynamics of each transfer.

\subsection{Reinforcement learning for spacecraft proximity operations}

Reinforcement Learning has been employed across a variety of rendezvous and proximity operations (RPO) tasks. For inspection specifically, \citet{vanwijk2023inspection} develop a low-level controller for a free-flying deputy to inspect a chief, incorporating considerations of solar illumination on the surface. \citet{aurand2025inspection} compare active (manoeuvring) and passive (natural-motion) inspection strategies, assessing fuel consumption and surface coverage within a multi-agent environment. In the context of collision avoidance, \citet{mu2024saferl} model debris avoidance involving a variable number of objects as a constrained Markov decision process (CMDP), solving it via a penalised policy-gradient method. \citet{bourriez2023cam} examine autonomous collision-avoidance decision-making under conditions of partial observability. These investigations exemplify the implementation of RL policies for proximity-operation tasks. However, many approaches operate at the level of continuous thrust commands and manage safety through reward shaping or fixed risk thresholds, typically resulting in a single operational point. Conversely, our approach functions at the discrete viewpoint selection level, validated by a low-level feasibility model, and explicitly incorporates safety as a tunable budget.

A parallel line of research emphasises safety through run-time assurance measures. \citet{vanwijk2024cbf} use discrete control barrier functions to filter the actions of an inspection agent. \citet{dunlap2024sixdof} extend this methodology to six degrees of freedom in inspection tasks, simultaneously enforcing constraints related to position, velocity, attitude, temperature, and power. \citet{dunlap2023rta} investigate centralised and decentralised assurance protocols for multi-agent inspection systems. \citet{hamilton2023ablation} present an ablation study examining the influence of the assurance mechanism on training outcomes and performance metrics. These filters certify safety based on an assumed model but do not specify a tolerated violation probability that a mission designer can establish. A chance constraint is employed to bound the residual risk remaining after accounting for navigation and manoeuvre execution errors, which is the regime addressed here. The two approaches act on different objects: one a deterministic envelope, the other a violation probability.

\subsection{Safe reinforcement learning and constrained MDPs}

The CMDP framework builds on the standard MDP by incorporating cost signals that need to stay below certain thresholds on average \citep{altman1999cmdp}. A broad survey of safe-RL methods and their guarantees is given by \citet{gu2024saferl}. Constrained Policy Optimization enforces a trust-region update that guarantees near-constraint satisfaction at every iteration \citep{achiam2017cpo}. Lagrangian methods approach the constraints by embedding them into the objective function with an adaptable multiplier. According to \citet{stooke2020pid}, treating this multiplier as part of a PID control loop damps oscillations and overshoot, leading to more stable constraint satisfaction during training. We use PPO \citep{schulman2017ppo} with a PID Lagrangian multiplier as our solver, since the constrained measure, the observed keep-out-violation rate, is an expected cumulative cost of the form the CMDP formalism treats.

\subsection{Invalid-action masking}

When certain actions are recognised as inadmissible within a specific state, invalid-action masking assigns their logits to negative infinity prior to the application of the policy's softmax function, thereby ensuring these actions are never sampled. \citet{huang2022masking} demonstrate that this masking technique corresponds to a legitimate policy-gradient update and that its advantages increase in proportion to the fraction of invalid actions. We employ masking for the constraints that follow deterministically from the nominal arc, specifically that a transfer permits a single-impulse CW solution within the designated per-manoeuvre $\Delta v$ budget. The keep-out clearance, subject to variation due to execution and navigation errors, is a random variable and is incorporated within the constrained policy.

\subsection{Relative motion model}

We model relative motion using the Clohessy--Wiltshire equations, a linearised description of a deputy's motion in the local-vertical/local-horizontal frame of a chief on a circular orbit \citep{clohessy1960,wakker2015}. The CW model admits closed-form impulsive transfers, which makes per-step feasibility and trajectory clearance inexpensive to evaluate and is therefore well suited to a planner that must screen many candidate manoeuvres on board. The model is a first-order approximation that assumes Newtonian point-mass gravity \citep{wakker2015}. We make this assumption explicit and revisit it when we discuss the validity of the static-frame approximation in Section~\ref{subsec:cw}.

\section{Problem Formulation and Solution Approach}
\label{section:method}

\subsection{Overview and architecture}

\begin{figure*}[tb]
\centering
\includegraphics[width=0.96\textwidth]{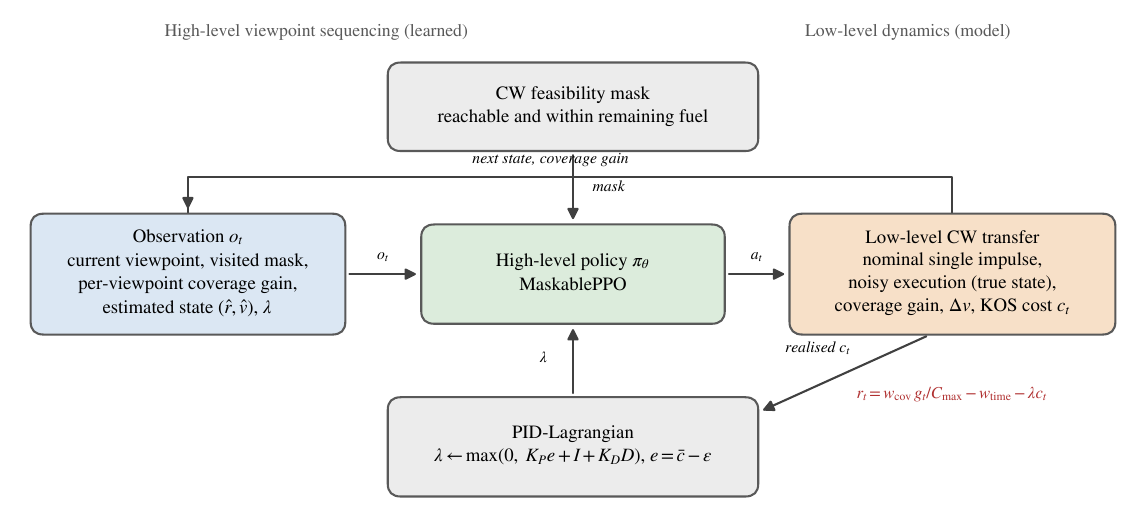}
\caption{Architecture. The high-level policy ($\pi_\theta$, MaskablePPO) chooses the next viewpoint from those allowed by the CW feasibility mask. The low-level model performs the transfer with execution and navigation noise, providing the coverage gain, fuel consumption, and actual keep-out cost $c_t$. A PID-Lagrangian multiplier $\lambda$, adjusted based on the rollout keep-out rate, influences the safety cost in the augmented reward $R_t=r_t-\lambda c_t$.}
\label{fig:arch}
\end{figure*}

We segment the inspection problem into two interconnected levels, as illustrated in Figure~\ref{fig:arch}. A \emph{high-level} policy determines, at each decision point, which of a predefined set of camera viewpoints to occupy subsequently. This decision constitutes the learned, sequential component of the task. A \emph{low-level} model validates and executes the commanded transfer using Clohessy--Wiltshire (CW) relative dynamics, provides the updated relative state, accounts for fuel consumption, and reports whether the realised trajectory has entered the designated keep-out sphere. The low-level module supplies the high-level with a feasibility mask, indicating viewpoints attainable within the per-manoeuvre and remaining fuel constraints, along with a keep-out cost signal. In turn, the high level directs the low level regarding the next viewpoint to reach. Under this architecture, kinematic reachability is deterministically derived from the nominal arc and is enforced by construction on the planned impulse, whereas keep-out clearance remains a stochastic variable under noisy execution and navigation conditions, and is subject to learning.

\subsection{Relative motion and single-impulse transfers}
\label{subsec:cw}

Relative motion is modelled within the target-centric local-vertical/local-horizontal (Hill) coordinate frame, where the $x$-axis points radially outward, the $y$-axis along-track, and the $z$-axis cross-track. For an object in a circular orbit with semi-major axis $a$ and mean motion $n=\sqrt{\mu/a^{3}}$, the linearised Clohessy--Wiltshire equations describe the deputy's relative position $\mathbf{r}=(x,y,z)$ \citep{clohessy1960,wakker2015}:
\begin{equation}
\ddot{x}-2n\dot{y}-3n^{2}x=0,\quad \ddot{y}+2n\dot{x}=0,\quad \ddot{z}+n^{2}z=0 .
\end{equation}

During coasting, the solution depends linearly on the initial state, $\mathbf{r}(t)=\Phi_{rr}(t)\,\mathbf{r}_0+\Phi_{rv}(t)\,\mathbf{v}_0$, with closed-form solutions available \citep{wakker2015}. A transfer from the current state $(\mathbf{r}_0,\mathbf{v}_0)$ to the target point $\mathbf{r}_d$ involves a single impulsive manoeuvre $\Delta\mathbf{v}$ at the start, followed by a coasting period lasting $\tau$. Ensuring the trajectory ends at the target yields
\begin{equation}
\Delta\mathbf{v}(\tau)=\Phi_{rv}(\tau)^{-1}\!\left[\mathbf{r}_d-\Phi_{rr}(\tau)\,\mathbf{r}_0\right]-\mathbf{v}_0 .
\label{eq:dv}
\end{equation}

The transfer is feasible if a coast time exists within the search window and the impulse respects the per-manoeuvre budget,
\begin{equation}
\exists\,\tau\in[\tau_{\min},\tau_{\max}]:\ \lVert\Delta\mathbf{v}(\tau)\rVert\le \Delta v_{\max}.
\label{eq:feas}
\end{equation}
The smallest such $\tau$ is chosen from a grid of $120$ candidate coast times within the search window. The safety margin is determined by the minimal clearance $\rho=\min_{t\in[0,\tau]}\lVert\mathbf{r}(t)\rVert$ of the arc from the target centre, which determines keep-out safety. The target attitude, and hence the viewpoint geometry, is held fixed within the rotating Hill frame over the inspection window, a quasi-static approximation supported by per-transfer coast times ($\tau_{\max}=1800$\,s) short relative to the ${\sim}14$-hour MEO period. Orbital perturbations are ignored for this reason. At this altitude, $J_2$ perturbation dominates, with differential $J_2$ accelerations between two spacecraft separated by tens of metres around $10^{-10}$\,m/s$^2$. This causes a displacement of less than a millimetre over the longest transfer, which is more than three orders of magnitude below the navigation errors already included in the model \citep{schweighart2002j2}.

\subsection{Inspection as a constrained Markov decision process}

We model viewpoint sequencing as a constrained Markov decision process (CMDP) \citep{altman1999cmdp}, consisting of a tuple $(\mathcal{S},\mathcal{A},P,r,C,\gamma)$ with reward $r$ and cost $C$. The target surface is represented by a point cloud of $N_p$ points. Each of the $N$ candidate viewpoints $p$ has a precomputed visibility set $\mathcal{V}_p$ of points within its field of view (FOV) that are not occluded (see Section~\ref{subsec:case}). The agent keeps track of the points it has already observed. The coverage state is defined by the visited subset, and the marginal gain of choosing a viewpoint is how many \emph{new} points it would reveal.

An action $a_t\in\{1,\dots,N\}$ involves selecting the next viewpoint.

The observation
\begin{equation}
o_t=\big[\,\mathbf{1}_{p_t};\ \mathbf{m}_t;\ \mathbf{g}_t;\ \hat{\mathbf{r}}_t,\hat{\mathbf{v}}_t;\ \lambda\,\big]\in\mathbb{R}^{3N+7},
\label{eq:obs}
\end{equation}
includes the one-hot encoding of the current viewpoint, a mask of visited viewpoints, the remaining marginal coverage gain for each viewpoint, the estimated relative state (Section~\ref{subsec:chance}), and the current multiplier. To keep the observation compact, individual view gains are used, since a full occupancy bitmap over the $N_p$ surface points would be roughly fifty times larger.

The objective is to maximise expected coverage while respecting a fuel budget and a keep-out constraint. Kinematic feasibility (\ref{eq:feas}) and the total $\Delta v$ budget are determined by the nominal arc and are enforced through action masking (Section~\ref{subsec:solver}). Since keep-out clearance is affected by execution and navigation errors, it is modelled as a random variable and incorporated into the learned constraint, which is described next.

\subsection{Stochastic execution and the keep-out chance constraint}
\label{subsec:chance}

We consider two sources of error: navigation and manoeuvre execution.

Navigation error occurs because the policy plans based on a noisy estimate $\hat{\mathbf{s}}_t=\mathbf{s}_t+\boldsymbol{\eta}$, $\boldsymbol{\eta}\sim\mathcal{N}(0,\Sigma_{\mathrm{nav}})$, of the relative state, while the actual state $\mathbf{s}_t$ determines the system's dynamics.

Execution error arises from an additive perturbation to the commanded impulse, $\Delta\tilde{\mathbf{v}}=\Delta\mathbf{v}+\boldsymbol{\xi}$ with $\boldsymbol{\xi}\sim\mathcal{N}(0,\sigma_{\Delta v}^{2}\,m^{2}\,\mathbf{I})$, a standard deviation proportional to the impulse magnitude, plus a minimum floor $m=\max(\lVert\Delta\mathbf{v}\rVert,\,0.01\,\mathrm{m/s})$ to prevent the perturbation from disappearing at small impulses.

The nominal impulse (\ref{eq:dv}) is calculated from the estimate $\hat{\mathbf{s}}_t$, but the actual arc is propagated from the true state $\mathbf{s}_t$ considering the perturbed impulse $\Delta\tilde{\mathbf{v}}$.

As a result, whether a transfer clears the keep-out sphere of radius $R_{\mathrm{KOS}}$ becomes probabilistic. We define the keep-out cost per step on the \emph{realised} arc,
\begin{equation}
c_t=\mathbf{1}\!\left[\rho^{\mathrm{real}}_t<R_{\mathrm{KOS}}\right],
\label{eq:cost}
\end{equation}
and apply a chance constraint on the expected violation rate over an episode length $T$, limited to a maximum $T_{\max}$,
\begin{equation}
J_C(\pi)=\mathbb{E}_\pi\!\left[\tfrac{1}{T}\textstyle\sum_{t} c_t\right]\le\varepsilon .
\label{eq:chance}
\end{equation}

\begin{figure*}[tb]
\centering
\includegraphics[width=0.92\textwidth]{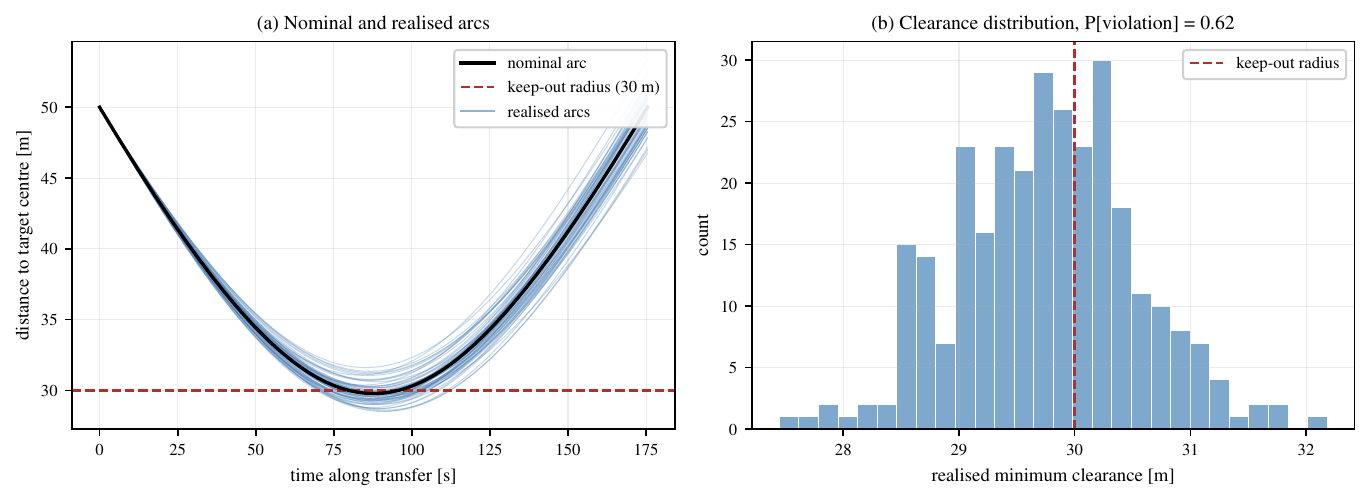}
\caption{The keep-out outcome under noise is inherently probabilistic. (a) In a grazing transfer, the nominal arc (black) reaches the keep-out radius, and the realised arcs (blue), drawn under execution noise, fall on both sides of it. (b) The distribution of the actual minimum-clearance values shows probability mass on both sides of $R_{\mathrm{KOS}}$, resulting in a violation probability of $0.62$ for this transfer. A deterministic test on the nominal arc returns a single pass or fail at this boundary and cannot express the violation probability the realised distribution carries.}
\label{fig:chance}
\end{figure*}

This constraint (\ref{eq:chance}) applies to the realised arc, where errors in navigation and execution influence risk. Figure~\ref{fig:chance} shows how this plays out for a grazing transfer, whose nominal clearance is $29.8$\,m, at the keep-out boundary. Under noise, the realised clearance can cross the boundary, resulting in a violation probability of $0.62$. Therefore, the agent must learn a safety \emph{margin} by choosing viewpoints that keep the violation rate below the allowed risk $\varepsilon$.

The single-step reward
\begin{equation}
r_t=w_{\mathrm{cov}}\,\frac{g_t}{C_{\max}}-w_{\mathrm{time}},
\label{eq:reward}
\end{equation}
rewards points newly covered relative to those that can be covered, minus a small per-step penalty. The keep-out cost $c_t$ affects the overall reward through the multiplier described in Section~\ref{subsec:solver}.

\subsection{Action masking and PID-Lagrangian PPO}
\label{subsec:solver}

Action masking involves setting the policy logits of inadmissible viewpoints to negative infinity before applying softmax, ensuring only valid actions are considered, which aligns with valid policy-gradient methods \citep{huang2022masking}. The admissible set is
\begin{equation}
\mathcal{A}_{\mathrm{feas}}(\hat{\mathbf{s}}_t)=\big\{\,p\neq p_t:\ \exists\,\tau,\ \lVert\Delta\mathbf{v}(\hat{\mathbf{s}}_t\!\to p,\tau)\rVert\le \min(\Delta v_{\max},\,b_t)\,\big\},
\label{eq:mask}
\end{equation}
with $b_t$ the remaining fuel. This masking enforces kinematic feasibility and the fuel limit on the planned impulse derived from the current estimate, so the multiplier influences only the keep-out risk. Actual consumption can differ from the planned impulse due to execution errors, typically by a few percent. If no admissible options remain, the episode terminates.

Lagrangian relaxation addresses the CMDP by transforming the constraint (\ref{eq:chance}) into the objective
\begin{equation}
\max_\theta\ \min_{\lambda\ge0}\ \ \mathcal{L}(\theta,\lambda)=\mathbb{E}_\pi\!\Big[\textstyle\sum_t \gamma^t r_t\Big]-\lambda\,(J_C-\varepsilon).
\label{eq:lagrangian}
\end{equation}
For a fixed multiplier, this results in an unconstrained problem in the augmented reward per step $R_t=r_t-\lambda\,c_t$. The policy is optimised using Proximal Policy Optimization \citep{schulman2017ppo}. The multiplier is adjusted with a PID scheme from \citet{stooke2020pid}, which reduces oscillations in the integral update. Constraint error $e_k=\bar{c}_k-\varepsilon$, measured as the average keep-out rate over a rollout $k$, is used in
\begin{align}
I_k&=\mathrm{clip}\!\left(I_{k-1}+K_I e_k,\ 0,\ I_{\max}\right),\quad D_k=\max(0,\,e_k-e_{k-1}),\nonumber\\
\lambda_k&=\max\!\left(0,\ K_P e_k+I_k+K_D D_k\right).
\label{eq:pid}
\end{align}

During training, the cost $c_t$ is evaluated on a \emph{single} realised trajectory per transfer, providing an unbiased sample of the violation event. The average $\bar c_k$ over the rollout gives the estimate for (\ref{eq:pid}). The risk budget $\varepsilon$ is a single tunable parameter that determines the operating point of the constrained policy.

\subsection{Case study and implementation}
\label{subsec:case}

\begin{figure*}[tb]
\centering
\includegraphics[width=0.96\textwidth]{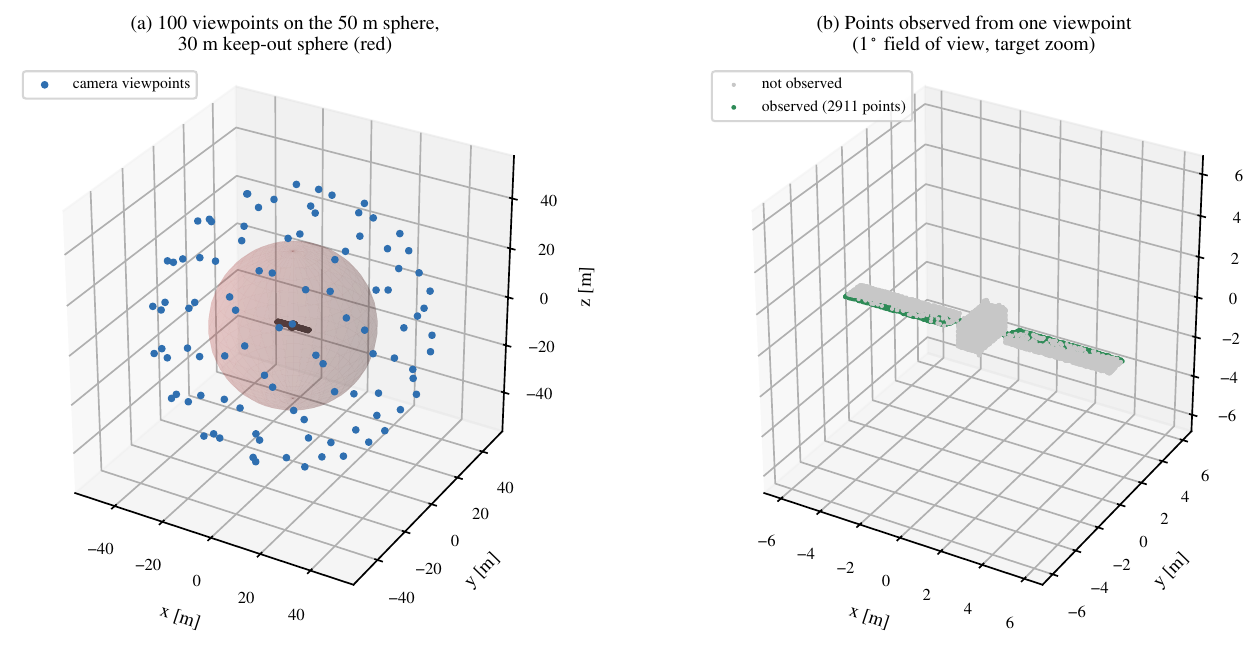}
\caption{Inspection case study. (a) The 100 camera viewpoints on the 50\,m sphere, the 30\,m keep-out sphere (red), and the Galileo point cloud at the centre. (b) Points observed from a single viewpoint within a $1^\circ$ field of view and subject to line-of-sight occlusion.}
\label{fig:geom}
\end{figure*}

We instantiate the problem on a Galileo-class navigation satellite in a circular medium Earth orbit at an altitude of $23{,}222$\,km. The target geometry comprises a $N_p=10{,}000$-point cloud representing the main bus and deployed solar panels, with a span of $13.0$\,m. One hundred camera viewpoints are positioned on a 50\,m sphere using a Fibonacci spiral to achieve near-uniform angular coverage \citep{hannay2004fibonacci}. From this range, the target subtends angles ranging from $3.0^\circ$ to $15.1^\circ$ (median $14.1^\circ$), depending on the viewing direction, thus exceeding the sensor cone at every viewpoint. Visibility is determined using a full-cone field of view of $1^\circ$ and a $z$-buffer line-of-sight occlusion test. Consequently, each viewpoint images a surface strip, which includes between $415$ and $2911$ points (median $624$). A 30\,m keep-out sphere encloses the target. Figure~\ref{fig:geom} illustrates the configuration and visibility from a single viewpoint. Coverage is expressed relative to the coverable set---the union of all viewpoints' visibility---which for this configuration is $74\%$ of the point cloud.

The policy and value functions are multilayer perceptrons comprising two hidden layers of 256 units each. They are trained using MaskablePPO from \texttt{sb3-contrib}, the contributed algorithms package of Stable-Baselines3 \citep{raffin2021sb3}, over $300{,}000$ environment steps, with reward normalisation. Table~\ref{tab:params} enumerates the environment and training parameters. To trace the coverage--risk frontier, a distinct policy is trained for each risk budget $\varepsilon$, with values set at $0.01$, $0.02$, $0.05$, $0.10$, and $0.20$. Three independent training seeds are employed for $\varepsilon$ values of $0.02$, $0.05$, and $0.20$. Each training run is evaluated over $1000$ independent episodes using a deterministic policy. Due to the stochastic nature of the environment, the episode-averaged keep-out rate provides a Monte Carlo estimate of $J_C$. An episode ends when the coverable set has been fully observed, when the step cap is reached, or when no admissible viewpoint remains.

Comparative analysis is conducted against heuristics evaluated under identical stochastic dynamics: a random feasible policy (uniform over the admissible set); a keep-out-blind coverage greedy policy (maximising marginal coverage over the admissible set without considering clearance); a robust-mask greedy policy (maximising marginal coverage among viewpoints whose nominal arcs clear the keep-out sphere by a 5\,m margin, i.e., at 35\,m); and a chance-constrained greedy policy (maximising marginal coverage among viewpoints whose estimated violation probability does not exceed its budget of $\varepsilon=0.05$, based on $K=12$ Monte Carlo samples per candidate). The greedy baselines use the dynamics model online to estimate per-viewpoint risk, whereas the learned policy employs this model solely during training. The chance-constrained greedy policy draws its Monte Carlo samples from the true relative state rather than from the estimate. It is therefore an oracle, bounding what a sampling planner could achieve given perfect navigation.

\begin{table}[t]
\centering
\caption{Environment and training parameters.}
\label{tab:params}
\footnotesize
\begin{tabular}{@{}lll@{}}
\toprule
Symbol & Description & Value \\
\midrule
$N$ & camera viewpoints & 100 \\
$N_p$ & point-cloud size & 10{,}000 \\
--- & viewpoint sphere radius & 50\,m \\
$R_{\mathrm{KOS}}$ & keep-out radius & 30\,m \\
--- & field of view & $1^\circ$ \\
$\Delta v_{\max}$ & per-manoeuvre budget & $0.5$\,m/s \\
--- & total $\Delta v$ budget & $20$\,m/s \\
$T_{\max}$ & max steps per episode & 50 \\
$\sigma_{\Delta v}$ & execution error (1$\sigma$) & 2\% \\
$\Sigma_{\mathrm{nav}}$ & navigation error (1$\sigma$) & $0.5$\,m,\ $5$\,mm/s \\
$\varepsilon$ & keep-out risk budget & $\{0.01\text{--}0.20\}$ \\
$w_{\mathrm{cov}},w_{\mathrm{time}}$ & reward weights & $10,\ 0.01$ \\
$K_P,K_I,K_D$ & PID gains & $0.1,\ 0.05,\ 0.01$ \\
$I_{\max}$ & integral anti-windup cap & 50 \\
$[\tau_{\min},\tau_{\max}]$ & coast-time search window & $[10,\ 1800]$\,s \\
--- & learning rate & $3\times10^{-4}$ \\
--- & rollout / batch / epochs & $1024 / 64 / 10$ \\
$\gamma$, $\lambda_{\mathrm{GAE}}$ & discount, advantage & $0.99,\ 0.95$ \\
\bottomrule
\end{tabular}
\end{table}

\section{Results}
\label{section:results}

Policies are assessed in accordance with the protocol and in comparison to the
reference policies outlined in Section~\ref{subsec:case}. We present the
\emph{coverage} achieved relative to the coverable set, the \emph{observed
keep-out violation rate}, the \emph{fuel} consumption $\sum_k |\Delta v_k|$ on
the actual trajectory, and the episode length in transfers. The violation
rate is defined as the proportion of executed transfers where the noisy arc
intersects the keep-out sphere; its mean across episodes serves as a Monte
Carlo estimate of the probability constrained by (\ref{eq:chance}).

\subsection{Training convergence}
\label{subsec:results-training}

\begin{figure}[t]
  \centering
  \includegraphics[width=0.9\columnwidth]{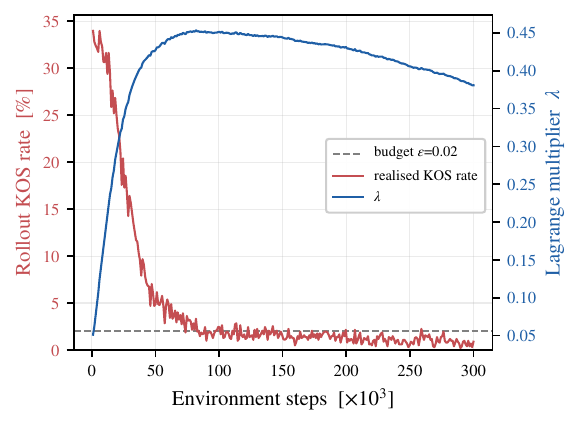}
  \caption{Training at $\varepsilon = 0.02$. The realised keep-out violation rate
  (left axis) falls below the budget as the PID multiplier $\lambda$ (right axis)
  increases, then relaxes to a steady value.}
  \label{fig:training}
\end{figure}

Figure~\ref{fig:training} illustrates a single training run at
$\varepsilon = 0.02$. The untrained policy breaches the keep-out sphere
approximately once in three transfers. The multiplier $\lambda$ increases as
this cost surpasses the allocated budget, overshooting to approximately $0.45$
near $80$k steps, before gradually decreasing as the integral term unwinds,
eventually stabilising near $0.38$. The rollout-mean violation rate decreases
below the $\varepsilon = 0.02$ threshold around $100$k steps, with its moving
average remaining below the budget thereafter. Coverage during the same period
rises to approximately $0.99$.

\subsection{Risk calibration and reference policies}
\label{subsec:results-frontier}

\begin{figure}[tb]
  \centering
  \includegraphics[width=0.85\columnwidth]{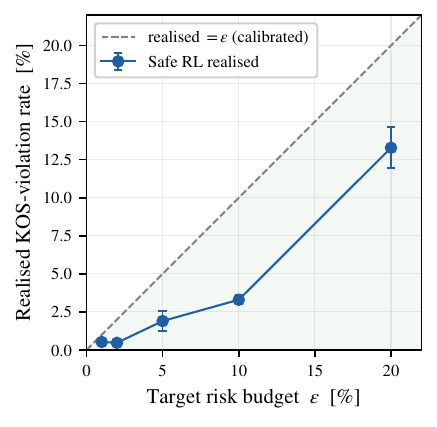}
  \caption{Realised keep-out violation rate against the requested budget
  $\varepsilon$. The realised rate lies on or below the dashed $y = x$ line at
  every tested budget. Error bars show one s.d.\ across three training seeds at
  $\varepsilon = 0.02$, $0.05$, $0.20$, and the standard error over $1000$
  episodes elsewhere.}
  \label{fig:calibration}
\end{figure}

Adjusting the budget $\varepsilon$ to levels of $0.01$, $0.02$, $0.05$, $0.10$,
and $0.20$ results in a policy whose realised violation rate aligns with the
specified threshold, as illustrated in Figure~\ref{fig:calibration}. The
observed rate remains at or below $\varepsilon$ within the examined range,
ranging from $0.5\%$ at the most restrictive budget to $13.3\%$ at the most
lenient. Across all tested budgets, a single design parameter determines the
operational risk without necessitating further retuning.
Analyses across three distinct training seeds at $\varepsilon$ values of
$0.02$, $0.05$, and $0.20$ reveal seed-to-seed variation in the realised rate
of $0.2$, $0.7$, and $1.4$ percentage points, respectively; in every instance,
the policy satisfies its budget, and the averaged rates across seeds increase
monotonically with $\varepsilon$. Both budgets with the highest restrictiveness
are associated with a realised rate of $0.5\%$.

An oracle greedy set cover method applied over the visibility sets achieves
coverage of $98.5\%$ of the coverable set using ten viewpoints, and all
safety-aware policies fall within the range of $97.8\%$ to $99.2\%$,
indicating that the risk budget does not impose a systematic coverage cost.
Moreover, tighter budgets coincide with episodes that are both less
costly and shorter in duration. As $\varepsilon$ decreases from $0.20$ to
$0.01$, fuel consumption decreases from $15.4$ to $10.6$~m/s, and the mean
episode length reduces from $32$ to $23$ transfers, while coverage variation
across budgets remains below one percentage point (refer to
Table~\ref{tab:results}). Achieving the most restrictive budget requires
neither additional fuel nor additional manoeuvres, given this specific geometry.

\begin{figure}[tb]
  \centering
  \includegraphics[width=\columnwidth]{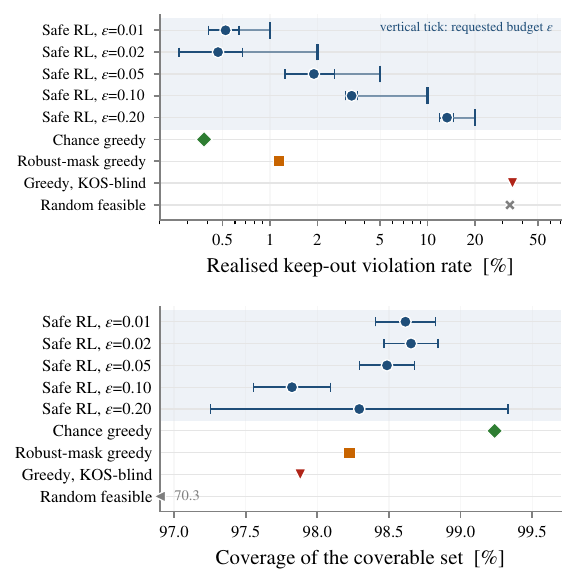}
  \caption{Policy comparison on a shared categorical axis. Upper: realised
  keep-out violation rate, logarithmic, with the vertical tick marking each
  learned policy's requested budget $\varepsilon$. Lower: coverage over a range
  chosen to resolve the safety-aware policies, with the random-feasible policy
  lying off scale at $70.3\%$. Error bars follow Figure~\ref{fig:calibration}.}
  \label{fig:frontier}
\end{figure}

\begin{table*}[tb]
\centering
\caption{Coverage, realised keep-out (KOS) violation rate, fuel and episode
length for the chance-constrained RL agent across risk budgets $\varepsilon$ and
the heuristic baselines. Coverage is reported relative to the coverable set.
All figures are Monte-Carlo means over 1000 episodes per run, and $n$ is the
number of independent training seeds averaged. Where $n = 3$, the stated
dispersion is one s.d.\ across seeds. Where $n = 1$, it is one s.d.\ across
episodes for coverage, and the remaining columns are single-run values. The
chance-greedy row is an oracle: its risk estimate is drawn from the true
relative state.}
\label{tab:results}
\begin{tabular}{lccccc}
\toprule
Policy & $n$ & Coverage [\%] & KOS rate [\%] & Fuel [m/s] & Steps \\
\midrule
Safe RL, $\varepsilon=0.01$ & 1 & 98.6 $\pm$ 6.6 & 0.52 & 10.6 & 23.2 \\
Safe RL, $\varepsilon=0.02$ & 3 & 98.7 $\pm$ 0.2 & 0.47 $\pm$ 0.20 & 11.5 & 24.7 \\
Safe RL, $\varepsilon=0.05$ & 3 & 98.5 $\pm$ 0.2 & 1.90 $\pm$ 0.65 & 13.0 & 27.7 \\
Safe RL, $\varepsilon=0.10$ & 1 & 97.8 $\pm$ 8.5 & 3.30 & 15.2 & 32.1 \\
Safe RL, $\varepsilon=0.20$ & 3 & 98.3 $\pm$ 1.0 & 13.28 $\pm$ 1.36 & 15.4 & 32.1 \\
\midrule
Random feasible & 1 & 70.3 & 33.28 & 18.0 & 37.2 \\
Greedy (KOS-blind) & 1 & 97.9 & 34.56 & 8.6 & 17.6 \\
Robust-mask greedy & 1 & 98.2 & 1.14 & 9.2 & 19.2 \\
Chance greedy & 1 & 99.2 & 0.38 & 9.3 & 19.3 \\
\bottomrule
\end{tabular}
\end{table*}

Figure~\ref{fig:frontier} and Table~\ref{tab:results} present the reference
policies under identical dynamic conditions. The application of feasibility
masking alone does not ensure safety. The randomly feasible policy breaches the
keep-out sphere in one-third of its transfers and achieves only $70\%$ coverage
of the set. The keep-out-blind greedy policy breaches the keep-out sphere in
$34.6\%$ of its transfers, as neither approach accounts for the realised
dynamics.

The remaining comparison delineates two approaches for addressing the keep-out
requirement. Applying a fixed margin to the keep-out radius and employing
deterministic planning, the method a mission designer would typically reach for
first, achieves a $98.2\%$ coverage rate with a $1.14\%$ violation rate.
When the requirement is considered as a chance constraint, the violation rate
falls sharply. At $\varepsilon = 0.02$, the learned policy attains $98.7\%$ coverage
with a $0.47\%$ violation rate, which is less than half the violation rate of
the fixed-margin heuristic. The oracle chance-constrained planner, sampling
from the true relative state, reaches $0.38\%$, so the learned policy operates
within $0.1$ percentage points of a bound that presumes perfect navigation. The
gain over the fixed margin stems from stating the constraint on the realised
arc, where the risk is concentrated.

The learned policy achieves this performance solely through reward and cost
signals, without requiring manually set keep-out margins or individual risk
assessments for each candidate. It exposes the tolerated violation probability
as an explicit parameter, which is set once and held across the sweep. The critical distinction on board is the cost associated with
decision-making. Each planner evaluated here assesses CW feasibility across
the candidate set. A sampling-based chance-constrained planner must
additionally generate $K$ noisy trajectories for each candidate, requiring the
stochastic model to be available onboard; in contrast, the learned policy
incorporates this estimation into its training process and only necessitates
the feasibility mask and a forward pass during execution.

\section{Conclusion}
\label{section:conclusions}

This paper models autonomous satellite inspection as viewpoint sequencing
within a chance-constrained Markov decision process. It enforces single-impulse
Clohessy--Wiltshire reachability and the fuel limit through invalid-action
masking. The keep-out constraint is a bound $\varepsilon$ on the violation
probability along the noise-affected trajectory. During training, a
PID-controlled Lagrange multiplier adjusts the penalty for this risk. For a
Galileo-class target in medium Earth orbit, the multiplier reduces the actual
violation rate below the set budget, and the trained policy remains within the
limit across a twentyfold $\varepsilon$ range, maintaining surface coverage
between $97.8\%$ and $98.7\%$. Achieving the strictest budget does not require
extra fuel or manoeuvres in this geometry, making safety a calibrated parameter
with no measured efficiency trade-off. Compared to a deterministic keep-out
heuristic, the chance-constrained policy attains less than half the violation
rate at comparable coverage. It also amortises the per-candidate risk estimate that
a chance-constrained heuristic repeats at each decision point, and this
approach can be adapted to other problem classes where viewpoint ordering
interacts with constraints.

These results carry two qualifications. The first concerns the coverage model.
The study rests on a single target geometry; the objective on that target
saturates within roughly ten viewpoints, so the ordering of viewpoints carries
less weight than it would on a larger or more heavily occluded structure;
visibility is a field-of-view and occlusion test without an incidence-angle
criterion; and coverage is evaluated at the commanded viewpoint, so quantifying
its sensitivity to arrival dispersion is a natural extension. The second
concerns the dynamics, which are linearised Clohessy--Wiltshire equations with
the target attitude held fixed in the rotating Hill frame, under Gaussian
navigation and execution errors. Future work will examine
regimes in which the fuel budget binds or the keep-out requirement couples
decisions across transfers, and will lift the quasi-static frame with
higher-fidelity relative dynamics.

\section*{Acknowledgments}
This work was supported by the \emph{Agence Nationale de la Recherche (ANR)}, France, under the project \textbf{ANR-23-CE10-0006} (Resilient and Sustainable Planning and Management of Future Space Industry Infrastructure with On-Orbit Servicing) for which Prof. Adam Abdin is the principal investigator. This work was supported by the Paris-Saclay Institute of Aeronautics and Astronautics (PSIA2).

\addcontentsline{toc}{section}{References}
\bibliographystyle{unsrtnat}
{
\renewcommand{\refname}{\normalsize References}
\footnotesize \bibliography{references}
}

\end{document}